\documentclass{article}
\usepackage{spconf}

\usepackage[english]{babel}

\usepackage{ifpdf}

\usepackage{cite} 
\usepackage{url}
\usepackage{hyperref}

\usepackage[pdftex]{graphicx}
\graphicspath{{./figures/}}
\usepackage{color}
\usepackage[table]{xcolor}
\usepackage{pgf, tikz, pgfplots}
\usetikzlibrary{shapes, arrows, automata, plotmarks}
\usetikzlibrary{calc,hobby,decorations}

\usepackage[cmex10]{amsmath}
\usepackage{amsfonts, amssymb, amsthm}
\usepackage{mathrsfs}

\usepackage{algorithm,algpseudocode}
\algnewcommand{\LeftComment}[1]{\Statex \(\triangleright\) #1}

\usepackage{enumerate}
\usepackage{multirow}
\usepackage{rotating}
\usepackage{subcaption}
\input{auxFiles/pennColors.sty}
\usepackage{needspace}

\newcommand{\myparagraph}[1]{\needspace{1\baselineskip}\medskip\noindent {\bf #1}}

\input{auxFiles/mySymbol.sty}

\def\myfactor{1.0}
\def\unit{\myfactor cm}

\pgfplotsset{ ylabel near ticks,                 
              xlabel near ticks,                 
              tick label style = {font=\footnotesize},   
              label style = {font=\footnotesize}, 
              title style = {font=\footnotesize},
            }

\tikzstyle{empty node} = [ circle, 
                     draw = black,
                     text = black, 
                     minimum size = 0.8*\unit]

\tikzstyle{node} = [ empty node, 
                     fill = blue!50,
                     draw = blue!50,
                     text = white]

\tikzstyle{blue node} = [ empty node, 
                         fill = blue!50,
                         draw = blue!50,
                         text = white]

\tikzstyle{red node} = [ empty node, 
                         fill = red!50,
                         draw = red!50,
                         text = white]

\tikzstyle{green node} = [ empty node, 
                         fill = mygreen!50,
                         draw = mygreen!50,
                         text = white]

\tikzstyle{black node} = [ empty node, 
                           fill = black!40,
                           draw = black!40,
                           text = white]

\tikzstyle{empty dot} = [ circle, 
                           draw = black,
                           inner sep = 0pt,
                           anchor = center,
                           minimum size = 0.4*\unit]

\tikzstyle{dot} = [ empty dot, 
                    fill = blue!50,
                    draw = blue!50]

\tikzstyle{blue dot} = [ empty dot, 
                         fill = blue!50,
                         draw = blue!50]

\tikzstyle{red dot} = [ empty dot, 
                        fill = red!50,
                        draw = red!50]

\tikzstyle{black dot} = [ empty dot, 
                          fill = black!50,
                          draw = black!50]

\tikzstyle{edge}                 = [shorten >=1pt, shorten <=1pt]
\tikzstyle{directed edge}        = [edge, -stealth]
\tikzstyle{double directed edge} = [edge, stealth-stealth]
\tikzstyle{tight edge}                 = [shorten >=0pt, shorten <=0pt]

\tikzstyle{block} = [ rectangle,
                      minimum width = \unit,
                      minimum height = \unit,
                      fill = blue!15,
                      draw = black,
                      text = black]

\newcommand{\cellpurple}{\cellcolor{pennpurple!15}}
\newcommand{\cellred}{\cellcolor{pennred!15}}
\newcommand{\cellorange}{\cellcolor{pennorange!15}}

\newcommand{\cellgreen}{\cellcolor{penngreen!15}}
\newcommand{\cellblue}{\cellcolor{pennblue!15}}

\title{GRAPH NEURAL NETWORKS FOR DECENTRALIZED CONTROLLERS}
\twoauthors
{Fer\hspace{0.015cm}nando Gama}
{University of California, Berkeley\\
    Electrical Eng. and Computer Sci. Dept.\\
    Berkeley, CA}
{Ekaterina Tolstaya, Alejandro Ribeiro\thanks{Supported by NSF CCF 1717120, ARO W911NF1710438, ARL DCIST CRA W911NF-17-2-0181, ISTC-WAS and Intel DevCloud.}}
{University of Pennsylvania\\
    Dept. of Electrical and Systems Engineering\\
    Philadelphia, PA}
\begin{document}
\ninept
\maketitle
\begin{abstract}
    Dynamical systems comprised of autonomous agents arise in many relevant problems such as multi-agent robotics, smart grids, or smart cities. Controlling these systems is of paramount importance to guarantee a successful deployment. Optimal \emph{centralized} controllers are readily available but face limitations in terms of scalability and practical implementation. Optimal \emph{decentralized} controllers, on the other hand, are difficult to find. In this paper, we propose a framework using graph neural networks (GNNs) to learn decentralized controllers from data. While GNNs are naturally distributed architectures, making them perfectly suited for the task, we adapt them to handle delayed communications as well. Furthermore, they are equivariant and stable, leading to good scalability and transferability properties. The problem of flocking is explored to illustrate the potential of GNNs in learning decentralized controllers.
\end{abstract}
%


\section{Introduction} \label{sec:intro}



Dynamical systems comprised of a set of autonomous agents arise in many technologically relevant scenarios. Examples include path planning in multi-agent robotics \cite{Li20-MultiPath}, optimal power allocation in smart grids \cite{Owerko20-OPF}, or traffic coordination in smart cities \cite{Li18-Traffic}. The ability to control network dynamical systems thus becomes a technological problem of paramount importance \cite{Sontag98-Control, Dorf08-Control, Gama19-Control}.

Optimal controllers for network systems have been obtained for a vast array of problems including constrained consensus in multi-agent systems \cite{Nedic10-Consensus}, load control in electrical grids \cite{Mohsenian10-LoadControl}, and throughput control of wireless networks \cite{Chiang07-PowerControl}. Computing these controllers, however, requires access to the state of the entire system at any time, rendering them \emph{centralized} solutions. Centralized controllers, albeit optimal, face limitations in terms of scalability and implementation.

Designing \emph{decentralized} controllers demands relying on the communication network established by the agents that compose the system. These agents can communicate only with other nearby agents and exchange information with them. A \emph{decentralized} controller, then, is built upon this distributed information structure. Furthermore, due to the inherent delay in the communication exchange, the information is not only distributed, but also outdated.

Optimal decentralized controllers are famously difficult to find \cite{Witsenhausen68-Counterexample}. Therefore, in this paper, we propose to \emph{learn} suitable controllers from data. More specifically, we propose a framework in which we adopt nonlinear maps between the local state and the control action given by two different types of graph neural networks (GNNs) \cite{Gama20-SPM}, either graph convolutional neural networks (GCNNs) \cite{Bruna14-DeepSpectralNetworks, Defferrard17-CNNGraphs, Kipf17-ClassifGCN, Gama19-Architectures} or graph recurrent neural networks (GRNNs) \cite{Ruiz20-GRNN}. While these learning models naturally respect the decentralized requirement of network systems, we further adapt them to handle delays.

GNNs are information processing architectures built upon the notion of graph filters \cite{Sandryhaila13-DSPG}, so as to exploit the graph structure, and the use of nonlinearities to increase expressive power \cite{Ruiz20-Nonlinear}. They are used to learn a nonlinear map between the state of the system and the action to be taken. In particular, GCNNs consist of a cascade of layers each of which applies a graph convolution \cite{Sandryhaila13-DSPG} followed by a pointwise nonlinearity \cite{Gama19-Architectures}. GRNNs, on the other hand, model the dynamical evolution of the network by learning a \emph{hidden} state that keeps track of this dynamic, and is then leveraged to compute the action. In this way, the hidden state learns to keep track of the relevant past information as it pertains to the actions taken \cite{Ruiz20-GRNN}.

Both GCNNs and GRNNs process only local information (relied by neighboring agents) and can be computed in a distributed manner (each agent computes its corresponding output). Furthermore, they both exhibit the properties of permutation equivariance and stability to perturbations \cite{Gama19-Stability, Ruiz20-GRNN}. The former means that these architectures exploit the topological symmetries in the underlying network, while the latter implies the output is not significantly affected by small changes in the network structure. These results allow GCNNs and GRNNs to scale up, i.e. to be used in networks of increasing number of agents. These results also imply that these learning models transfer, i.e. they can be trained in one network and then transferred to another similar network.

Scalability and transferability become crucial properties when we train the models via the \emph{imitation learning} framework \cite{Ross10-ImitationLearning, Ross11-DAGger}. In this framework, a training set comprised of optimal trajectories is available and the models learn to \emph{imitate} these trajectories. The imitation learning framework rests on the availability of an optimal centralized controller which is only required during the offline training phase but is not needed during testing. For this reason, training architectures that scale and transfer is the key to learning decentralized controllers that are successful in previously unseen testing scenarios.

To illustrate the power of GCNNs and GRNNs in learning decentralized controllers, we explore the problem of flocking \cite{Tanner04-Flocking, Tolstaya19-Flocking}. The objective is to coordinate a team of agents, initially flying at random velocities, to fly at the same velocity while avoiding collisions. An optimal centralized solution is readily available \cite{Tanner03-Stable} which makes this problem well suited for the imitation learning framework.

In Sec.~\ref{sec:decentralizedControllers} we formulate the problem of learning decentralized controllers and introduce the framework of imitation learning. In Sec.~\ref{sec:GNN} we present the GNN and GRNN as learning architectures. In Sec.~\ref{sec:flocking} we discuss the problem of flocking. We conclude in Sec.~\ref{sec:conclusions}.


\section{Decentralized Controllers} \label{sec:decentralizedControllers}



Consider a set of $N$ agents $\ccalV = \{1, \ldots, N\}$. At time $t=0,1,2,\ldots$, each agent is described by a state vector $\bbx_{i}(t) \in \reals^{F}$ and is capable of taking an action $\bbu_{i}(t) \in \reals^{G}$. The time index $t$ represents the sequence of instances at sampling time $T_{s}$. The collection of states for all agents can be conveniently described by a $N \times F$ matrix $\bbX(t)$ where row $i$ corresponds to the state $\bbx_{i}(t)$ of agent $i$. Likewise, $\bbU(t)$ is the $N \times G$ matrix that collects the actions of all agents. The dynamic evolution of the system is given by some function $\ccalD$ that takes the past states and actions and outputs the current state $\bbX(t+1) = \ccalD(\{\bbX(\tau), \bbU(\tau)\}_{\tau = 0,\ldots,t})$. This function determines how the actions shape the future states.

The optimal actions to take $\bbU^{\star}(t)$ are given by those that minimize some cost function $\ccalJ$ over time
\begin{equation} \label{eqn:optimalAction}
    \{\bbU^{\star}(t)\}_{t} = \argmin_{\bbU(t), t \geq 0} \sum_{t} \ccalJ \big[ \bbX(t) \big].
\end{equation}
The optimal actions typically rely on the the state of the entire network, making them a \emph{centralized} optimal solution; they require centralized knowledge of the states of all agents in order to make a decision. We are concerned, however, with actions that are \emph{decentralized}. This means that (i) actions rely on information provided by neighboring agents only, and (ii) actions can be computed individually by each agent. Furthermore, we are interested in actions that respect the delayed nature of the communications.

The communication capability of the agents defines a dynamic communication network. This network can be conveniently described by means of a graph $\ccalG(t) = (\ccalV, \ccalE(t), \ccalW(t))$ where $\ccalE(t) \subseteq \ccalV \times \ccalV$ is the set of edges and $\ccalW(t): \ccalE(t) \to \reals_{+}$ is the weight function. Agents $i$ and $j$ can communicate at time $t$ if and only if $(i,j) \in \ccalE(t)$, and the weight function $\ccalW(t)$ can be used to describe the state of the channel. The neighborhood $\ccalN_{i}(t)$ of agent $i$ corresponds of all the agents to whom agent $i$ can communicate at time $t$, $\ccalN_{i}(t) = \{ j \in \ccalV : (j,i) \in \ccalE(t) \}$. A matrix description $\bbS(t)$ can be associated to graph $\ccalG(t)$. Matrix $\bbS(t)$ respects the sparsity of the graph, $[\bbS(t)]_{ij} = s_{ij}(t) = 0$ whenever $(j,i) \notin \ccalE(t)$ for $j \neq i$. Examples of such a matrix include the adjacency \cite{Sandryhaila13-DSPG}, the Laplacian \cite{Shuman13-SPG} or the Markov matrix \cite{Heimowitz17-MarkovGSP}, among others. In what follows we denote it, generically, the \emph{graph shift operator} \cite{Ortega18-GSP}.

Agents can communicate their states $\bbx_{i}(t)$ to other neighboring agents by means of this communication network. The transmission of information incurs in a unit delay, creating a \emph{partial information structure}
\begin{equation} \label{eqn:partialInformation}
    \ccalX_{i}(t) = \bigcup_{k=0}^{K-1} \Big\{ \bbx_{j}(t-k) : j \in \ccalN_{i}^{k}(t) \Big\}
\end{equation}
where $\ccalN_{i}^{k}(t)$ is the set of nodes $k$ hops away from node $i$ and is defined recursively as  $\ccalN_{i}^{k}(t) = \{ j' \in \ccalN_{j}^{k-1}(t-1) : j \in \ccalN_{i}(t)\}$ with $\ccalN_{i}^{1}(t) = \ccalN_{i}(t)$ and $\ccalN_{i}^{0}(t) = \{i\}$. We denote by $\ccalX(t) = \{\ccalX_{i}(t)\}_{i=1,\ldots,N}$ the collection of the \emph{information history} $\ccalX_{i}(t)$ of all nodes. The objective of this paper is to obtain controllers that respect this decentralized and delayed information structure $\bbU(t) = \bbPhi(\ccalX(t))$ as opposed to centralized controllers that take into account the information of all agents $\bbU(t) = \bbPsi(\{\bbX(t)\})$.

We consider a data-driven approach based on the framework of \emph{imitation learning} \cite{Ross10-ImitationLearning}. In this setting, a training set of optimal trajectories $\ccalT = \{(\bbX(t), \bbU^{\star}(t))_{t}\}$ is available, and the decentralized controller $\bbPhi(\ccalX(t))$ is found by tracking the optimal solution $\bbU^{\star}(t)$
\begin{equation} \label{eqn:imitationLearning}
    \min_{\bbPhi} \sum_{\ccalT} \big\| \bbPhi(\ccalX(t)) - \bbU^{\star}(t) \big\|.
\end{equation}
Imitation learning requires the computation of the optimal centralized action $\bbU^{\star}(t)$ [cf. \eqref{eqn:optimalAction}] but only for the training phase, i.e. an offline phase where \eqref{eqn:imitationLearning} is solved. Nevertheless, centralized controllers might be computationally expensive or available only for small networks. Therefore, for the learned map $\bbPhi$ to be useful, we need it to be scalable and computationally efficient. Additionally, solving \eqref{eqn:imitationLearning} over the space of all functions that operate on the partial information structure $\ccalX(t)$, are scalable and are efficient, is computationally intractable. Therefore, in what follows, we adopt a parametrization in terms of GCNNs or GRNNs. They not only satisfy the partial information structure \eqref{eqn:partialInformation}, but are also scalable and have an efficient, distributed implementation, as shown next.


\section{Graph Neural Networks} \label{sec:GNN}



A GNN is an information processing architecture built upon the notion of graph filters and the use of nonlinearities \cite{Gama20-SPM, Isufi20-EdgeNets}. In particular, GCNNs \cite{Gama19-Architectures} and GRNNs \cite{Ruiz20-GRNN} exploit the operation of graph convolution \cite{Sandryhaila13-DSPG} and use pointwise nonlinearities, resulting in local architectures that only involve communication with nearby agents, thus respecting the partial information structure. Additionally, they have a naturally distributed implementation, and are also stable and equivariant, helping in transfer learning and scalability \cite{Gama19-Stability, Ruiz20-GRNN}.

\myparagraph{Graph Convolutions.} The basic building block of both the GCNNs and the GRNNs are graph convolutions \cite{Sandryhaila13-DSPG, Gama19-GraphConv}. In analogy to time convolutions, a graph convolution is defined as a linear combination of shifted versions of the signal. The notion of \emph{shift} follows from applying the shift operator $\bbS(t)$ to the states $\bbX(t)$ whereby the $(i,f)$ element of the multiplication $\bbS(t)\bbX(t)$ is given by
\begin{equation} \label{eqn:graphShift}
    [\bbS(t) \bbX(t)]_{if} = \sum_{j=1}^{N} [\bbS(t)]_{ij}[\bbX(t)]_{jf} = \sum_{j \in \ccalN_{i}(t)} s_{ij}(t) x_{jf}(t).
\end{equation}
Due to the sparsity pattern of $\bbS(t)$, the shifting operation $\bbS(t) \bbX(t)$ is a linear combination of the values of the $f$th entry of the state in neighboring nodes only. This renders $\bbS(t)\bbX(t)$ a local and distributed operation, i.e. it needs only neighboring information, and each agent can compute the output separately.

Equipped with the notion of shift, we formally define the graph convolution as $\bbU(t) = \ccalA \ast_{\bbS(t)} \bbX(t)$ \cite{Sandryhaila13-DSPG, Gama20-SPM, Gama19-GraphConv}, where
\begin{equation} \label{eqn:graphConv}
    \ccalA \ast_{\bbS(t)} \bbX(t) \! = \! \sum_{k=0}^{K-1} \bbS(t) \bbS(t-1) \cdots \bbS(t-(k-1)) \ \bbX(t-k) \ \bbA_{k}
\end{equation}
with $\ccalA = \{\bbA_{k}\}_{k}$ being the convolution coefficients (filter taps or weights). Repeated application of delayed versions of the shift gathers information from further away neighbors [cf. \eqref{eqn:graphShift}]. In fact, the $k$th summand in \eqref{eqn:graphConv} gathers the correspondingly delayed information from nodes located in the $k$-hop neighborhood. This information is accessed by $k$ successive communication exchanges with the immediate neighbors. The output $\bbU(t)$ collects in its rows the resulting states $\bbu_{i}(t) \in \reals^{G}$ at each of the agents and need not have the same dimension $F$ as the input. As a matter of fact, the matrix of coefficients $\bbA_{k}$ is a $F \times G$ rectangular matrix that mixes the features located at each node. The graph convolution \eqref{eqn:graphConv} can be understood as the application of a $F\times G$ filter bank, with each matrix of coefficients $\bbA_{k}$ assigning weight to the information located in the $k$-hop neighborhood. From \eqref{eqn:graphShift} it follows that the graph convolution is a local and distributed linear operation.

\myparagraph{Graph Convolutional Neural Networks (GCNNs).} Graph convolutions are linear mappings between the input $\bbX(t)$ and the output $\bbU(t)$. Nonlinear behaviors can be described by adding a pointwise nonlinearity at the output of the graph convolution. A single-layer GCNN is obtained as \cite{Gama19-Architectures}
\begin{equation} \label{eqn:GNN}
    \bbU(t) = \sigma \big( \ccalA \ast_{\bbS(t)} \bbX(t) \big)
\end{equation}
where $\sigma$ is a nonlinearity that acts entrywise on the output of the convolution. Since the nonlinearity $\sigma$ is pointwise, then the GCNN inherits the properties of locality and distributed implementation from graph convolutions. Furthermore, GCNNs are permutation equivariant and stable to graph perturbations \cite{Gama19-Stability}. The former implies that the GCNN adequately exploit graph topological symmetries to enhance learning, while the latter means that the GCNN can transfer to other graph supports as long as the graphs are similar. As a matter of fact, note that once the coefficients $\ccalA$ are learned, they can be used on any graph shift operator $\bbS(t)$ [cf. \eqref{eqn:graphConv}]. Together, equivariance and stability confirm that GCNNs scale to larger graphs.

\myparagraph{Graph Recurrent Neural Networks (GRNNs).} To further increase the descriptive power of the parametrized mapping between the agent states $\bbX(t)$ and the actions taken $\bbU(t)$, we consider GRNNs \cite{Ruiz20-GRNN}. GRNNs learn a hidden state $\bbZ(t)$ from the sequence $\{\bbX(t)\}$ as follows
\begin{equation} \label{eqn:GRNNhidden}
    \bbZ(t) = \sigma \big( \ccalA \ast_{\bbS(t)} \bbX(t) + \ccalB \ast_{\bbS(t)} \bbZ(t-1)\big)
\end{equation}
where $\ccalA = \{\bbA_{k}\}_{k}$ and $\ccalB = \{\bbB_{k} \}_{k}$ contain the matrix of coefficients for the input-to-hidden and hidden-to-hidden convolutions. The hidden state matrix $\bbZ(t)$ collects in its rows the hidden state of each agent $\bbz_{i}(t) \in \reals^{H}$. The matrices of coefficients are then $\bbA_{k} \in \reals^{F \times H}$ and $\bbB_{k} \in \reals^{H \times H}$ for all $k$. The hidden state is then mapped into the learned action by means of another graph convolution followed by a pointwise nonlinearity
\begin{equation} \label{eqn:GRNNoutput}
    \bbU(t) = \rho \big( \bbC \ast_{\bbS(t)} \bbZ(t) \big)
\end{equation}
with $\rho$ another pointwise nonlinearity and $\ccalC = \{\bbC_{k}\}_{k}$ the collection of matrix of coefficients $\bbC_{k} \in \reals^{H \times G}$. Training a GRNN entails jointly optimizing \eqref{eqn:GRNNhidden}-\eqref{eqn:GRNNoutput}. Thus, the hidden state learns to keep track of the relevant past information as it pertains to the target output. As it follows from \eqref{eqn:graphConv}, GRNNs are also local and distributed. We note that they are also permutation equivariant and stable, allowing them to scale and transfer \cite{Ruiz20-GRNN}.

\medskip

It is crucial to note that all architectures introduced in this section, namely the graph convolutions \eqref{eqn:graphConv}, the GCNNs \eqref{eqn:GNN} and the GRNNs \eqref{eqn:GRNNhidden}-\eqref{eqn:GRNNoutput}, satisfy the partial information structure \eqref{eqn:partialInformation}. This implies that all of these are suitable parametrizations $\bbPhi(\ccalX(t))$ to learn \emph{decentralized} controllers.


\section{Flocking} \label{sec:flocking}



To illustrate the power of GCNNs and GRNNs in learning decentralized controllers, we explore the problem of flocking. The objective is to coordinate a team of agents, initially flying at random velocities, to fly at the same velocity while avoiding collisions. 

\myparagraph{Problem statement.} Each agent $i \in \ccalV$ is described by its position $\bbr_{i}(t) \in \reals^{2}$, its velocity $\bbv_{i}(t) \in \reals^{2}$ and its acceleration $\bbu_{i}(t) \in \reals^{2}$. The dynamic evolution $\ccalD$ of the system is given by
\begin{equation} \label{eqn:systemDynamics}
    \begin{aligned}
        \bbr_{i}(t+1) &= \bbu_{i}(t) T_{s}^{2}/2 + \bbv_{i}(t) \bbT_{s} + \bbr_{i}(t) \\
        \bbv_{i}(t+1) &= \bbu_{i}(t) T_{s} + \bbv_{i}(t)
    \end{aligned}
\end{equation}
for $i=1,\ldots,N$. These dynamics imply that the acceleration $\bbu_{i}(t)$ is held constant for the duration of the sampling interval $[tT_{s},(t+1)T_{s})$. The acceleration $\bbu_{i}(t)$ is the actionable variable, and we assume that the agents can adjust it instantaneously between sampling intervals. Formally, the objective of flocking is to determine the accelerations $\{\bbU(t)\}_{t}$ that make the velocities of all agents in the team be the same. This can be written as $\min_{\bbU(t), t \geq 0} \sum_{t}\ccalJ [\bbV(t)]$ [cf. \eqref{eqn:optimalAction}] with cost function
\begin{equation} \label{eqn:flockingObjective}
    \ccalJ\big[ \bbV(t) \big] =  \frac{1}{N} \sum_{i=1}^{N} \Big\| \bbv_{i}(t) - \frac{1}{N} \sum_{j=1}^{N} \sum_{j=1}^{N} \bbv_{j}(t) \Big\|^{2}
\end{equation}
subject to the system dynamics \eqref{eqn:systemDynamics}.

\myparagraph{Optimal centralized solution.} The optimal \emph{centralized} solution, while avoiding collisions, is given by accelerations $\bbU^{\star}(t)$, where the $i$th row is computed as
\begin{equation} \label{eqn:optimalActionFlocking}
    \bbu_{i}^{\star}(t) = - \sum_{j=1}^{N} \Big( \bbv_{i}(t) - \bbv_{j}(t) \Big) - \sum_{j=1}^{N} \nabla_{\bbr_{i}(t)} U \Big( \bbr_{i}(t), \bbr_{j}(t)\Big)
\end{equation}
where
\begin{align} \label{eqn:collisionAvoidance}
    & U\Big( \bbr_{i}(t), \bbr_{j}(t) \Big) \\
    & \quad =
    \begin{cases}
        1/\|\bbr_{ij}(t)\|^{2} - \log(\|\bbr_{ij}(t)\|^{2}) & \text{if} \|\bbr_{ij}(t) \| \leq \rho \\
        1/\rho^{2} - \log(\rho^{2}) & \text{otherwise}
    \end{cases} \nonumber
\end{align}
is the collision avoidance potential, with $\bbr_{ij}(t) = \bbr_{i}(t) - \bbr_{j}(t)$. Certainly, $\bbu^{\star}_{i}(t)$ is a centralized controller since computing it requires agent $i$ to have instantaneous knowledge of the velocity and position of every other agent in the team.

\myparagraph{Communication capabilities.} The communication network between agents is determined by their proximity. If agents $i$ and $j$ are within a communication radius $R$ of each other then they are able to establish a link. This builds a communication graph with edge set $\ccalE(t)$ such that $(i,j) \in \ccalE(t)$ if and only if $\|\bbr_{i}(t) - \bbr_{j}(t) \| \leq R$. The weight function $\ccalW(t)$ is $1$ for $(i,j) \in \ccalE(t)$ and $0$ otherwise, yielding a binary adjacency matrix that we adopt as the shift operator $\bbS(t)$. We assume that communication exchanges occur within the interval determined by the sampling time $T_{s}$, so that the action clock and the communication clock coincide.

\myparagraph{Learning decentralized controllers.} The communication network imposes a partial information structure $\ccalX(t)$ [cf. \eqref{eqn:partialInformation}]. We aim at learning controllers $\bbPhi(\ccalX(t))$ that respect this structure and to do so, we parametrize them following the models introduced in Sec.~\ref{sec:GNN}. We consider graph convolutions $\bbPhi_{\text{GC}}(\bbX(t); \bbS(t), \ccalA)$ given by \eqref{eqn:graphConv}, GCNNs $\bbPhi_{\text{GCNN}}(\bbX(t); \bbS(t), \ccalA)$ and GRNNs $\bbPhi_{\text{GRNN}}(\bbX(t); \bbS(t),$ $\ccalA,$ $\ccalB,$ $\ccalC)$. The notation emphasizes the partial information structure $\ccalX(t)$ described in terms of $\{\bbX(t)\}_{t}$ and $\{\bbS(t)\}_{t}$ and the collection of learnable coefficients $\ccalA$, $\ccalB$ or $\ccalC$, as appropriate. The agent states $\bbx_{i}(t) \in \reals^{6}$ that we consider for the flocking problem are given by
\begin{equation} \label{eqn:flockingState}
    \begin{aligned}
        \bbx_{i}(t) = \bigg[ & \sum_{j \in \ccalN_{i}(t)} \big(\bbv_{i}(t) - \bbv_{j}(t) \big), \\
        & \sum_{j \in \ccalN_{i}(t)} \frac{\bbr_{ij}(t)}{\| \bbr_{ij}(t)\|^{4}} , \sum_{j \in \ccalN_{i}(t)} \frac{\bbr_{ij}(t)}{\| \bbr_{ij}(t)\|^{2}} \bigg].
    \end{aligned}
\end{equation}
We exploit the imitation learning framework, simulating trajectories with the optimal centralized controller given by \eqref{eqn:optimalActionFlocking} and solving \eqref{eqn:imitationLearning} for each of the parametrizations.

%
\begin{table*}[ht]
    \scriptsize
    \centering
    \caption{Average (std. deviation) cost for different hyperparameters in flocking for all tested architectures. Optimal cost: $52(\pm 1)$.}
    \begin{subtable}[t]{0.39\textwidth}
        \centering
        \begin{tabular}{c|ccc} 
            $G$ / $K$ & $2$ & $3$ & $4$  \\ \hline
            $16$ & \cellpurple $521 (\pm 90)$ & \cellred $434 (\pm 78)$ & \cellred $404 (\pm 46)$ \\ 
            $32$ & \cellpurple $593 (\pm 175)$ & \cellred $433 (\pm  54)$ & \cellred $\mathbf{345 (\pm 37)}$ \\
            $64$ & \cellpurple $508 (\pm 96)$ & \cellred $419 (\pm  52)$ & \cellred $401 (\pm 56)$ \\ 
        \end{tabular}
        \caption{\small $\bbPhi_{\text{GC}}$}
        \label{tab:hParamGC}
    \end{subtable}
    \begin{subtable}[t]{0.3\textwidth}
        \centering
        \begin{tabular}{ccc} 
            $2$ & $3$ & $4$  \\ \hline
            \cellorange $177 (\pm 15)$ & \cellorange $169 (\pm 10)$ & \cellorange $162 (\pm 13)$ \\ 
            \cellblue $98 (\pm 3)$ & \cellblue $96 (\pm  5)$ & \cellblue $94 (\pm 3)$ \\ 
            \cellgreen $86 (\pm 9)$ & \cellgreen $\mathbf{83 (\pm  4)}$ & \cellgreen $85 (\pm 4)$ \\ 
        \end{tabular}
        \caption{\small $\bbPhi_{\text{GCNN}}$}
        \label{tab:hParamGNN}
    \end{subtable}
    \begin{subtable}[t]{0.3\textwidth}
        \centering
        \begin{tabular}{ccc} 
            $2$ & $3$ & $4$  \\ \hline
            \cellorange $171 (\pm 7)$ & \cellorange $169 (\pm 6)$ & \cellorange $161 (\pm 6)$ \\
            \cellblue $100 (\pm 7)$ & \cellblue $94 (\pm  4)$ & \cellblue $96 (\pm 3)$ \\
            \cellgreen $83 (\pm 2)$ & \cellgreen $90 (\pm  15)$ & \cellgreen $\mathbf{82 (\pm 6)}$ \\
        \end{tabular}
        \caption{\small $\bbPhi_{\text{DAGNN}}$}
        \label{tab:hParamDAGNN}
    \end{subtable}
    \begin{subtable}[t]{0.39\textwidth}
        \centering
        \begin{tabular}{c|ccc} 
            $G$ / $K$ & $2$ & $3$ & $4$  \\ \hline
            $16$ & \cellorange $140 (\pm 8)$ & \cellorange $133 (\pm 7)$ & \cellorange $135 (\pm 5)$ \\
            $32$ & \cellblue $83 (\pm 3)$ & \cellblue $82 (\pm  3)$ & \cellblue $82 (\pm 3)$ \\ 
            $64$ & \cellgreen $77 (\pm 2)$ & \cellgreen $\mathbf{77 (\pm  2)}$ & \cellgreen $77 (\pm 3)$ \\ 
        \end{tabular}
        \caption{\small $\bbPhi_{\text{GRNN}}$}
        \label{tab:hParamGRNN}
    \end{subtable}
    \begin{subtable}[t]{0.6\textwidth}
        \centering
        \begin{tabular}{c|ccccc} 
            $N$ & $50$ & $62$ & $75$ & $87$ & $100$  \\ \hline
            $\bbPhi_{\text{GC}}$ & $440(\pm 63)$ &  $406 (\pm 88)$ & $456 (\pm 69)$ & $459 (\pm 56)$ & $472(\pm 77)$ \\ 
            $\bbPhi_{\text{GCNN}}$ & $91 (\pm 8)$ & $92 (\pm 8)$ & $91 (\pm 5)$ & $101 (\pm 16)$ & $94 (\pm 11)$ \\ 
            $\bbPhi_{\text{DAGNN}}$ & $89 (\pm 8)$ & $89 (\pm 16)$ & $89 (\pm 8)$ & $88 (\pm 7)$ & $91 (\pm 15)$ \\
            $\bbPhi_{\text{GRNN}}$ & $78 (\pm 3)$ & $76 (\pm 2)$ & $77 (\pm 2)$ & $77 (\pm 2)$ & $77 (\pm 2)$ \\ 
        \end{tabular}
        \caption{Architectures trained on $50$ agents, and tested on $N$ agents.}
        \label{tab:scalability}
    \end{subtable}
    \label{tab:flocking}
\end{table*}
%

\myparagraph{Parametrizations.} For all three parametrizations we consider a single layer architecture with $G$ output features, $H=G$ hidden state features and $K$ filter taps. The nonlinearities are $\tanh$. In all cases, there is a local readout layer that maps the output $G$ features at each node into the $2$ acceleration components $\hbu_{i}(t)$. We compare these three parametrizations with the delayed aggregation GNN $\bbPhi_{\text{DAGNN}}$ \cite{Gama19-Aggregation}, which is the same parametrization used in \cite{Tolstaya19-Flocking}. 

\myparagraph{Dataset.} The dataset is comprised of $400$ trajectories for training, $20$ for validation and $20$ for testing. Each trajectory is generated by positioning the $N=50$ agents at random in a circle such that their minimum initial distance is $0.1\text{m}$ and their initial velocities are picked also at random from the interval $[-3,3]\text{m}/\text{s}$ in each direction, with a random bias for all agents selected from the same interval. The trajectories last $2\text{s}$ with sampling $T_{s} = 0.01\text{s}$, the maximum acceleration is $10\text{m}/\text{s}^{2}$ and the communication radius is $R=2\text{m}$.

\myparagraph{Training and evaluation.} The architectures are trained for $30$ epochs with a batch size of $20$ trajectories, following the ADAM optimizer with learning rate $5 \cdot 10^{-4}$ and forgetting factors $0.9$ and $0.999$. The evaluation measure is the cost \eqref{eqn:flockingObjective}. We repeat the simulations for $5$ realizations of the dataset and report the average cost as well as the standard deviation.

%
\begin{figure}[b]
    \centering
    \begin{subfigure}{0.495\columnwidth}
        \centering
        \includegraphics[width = \textwidth]{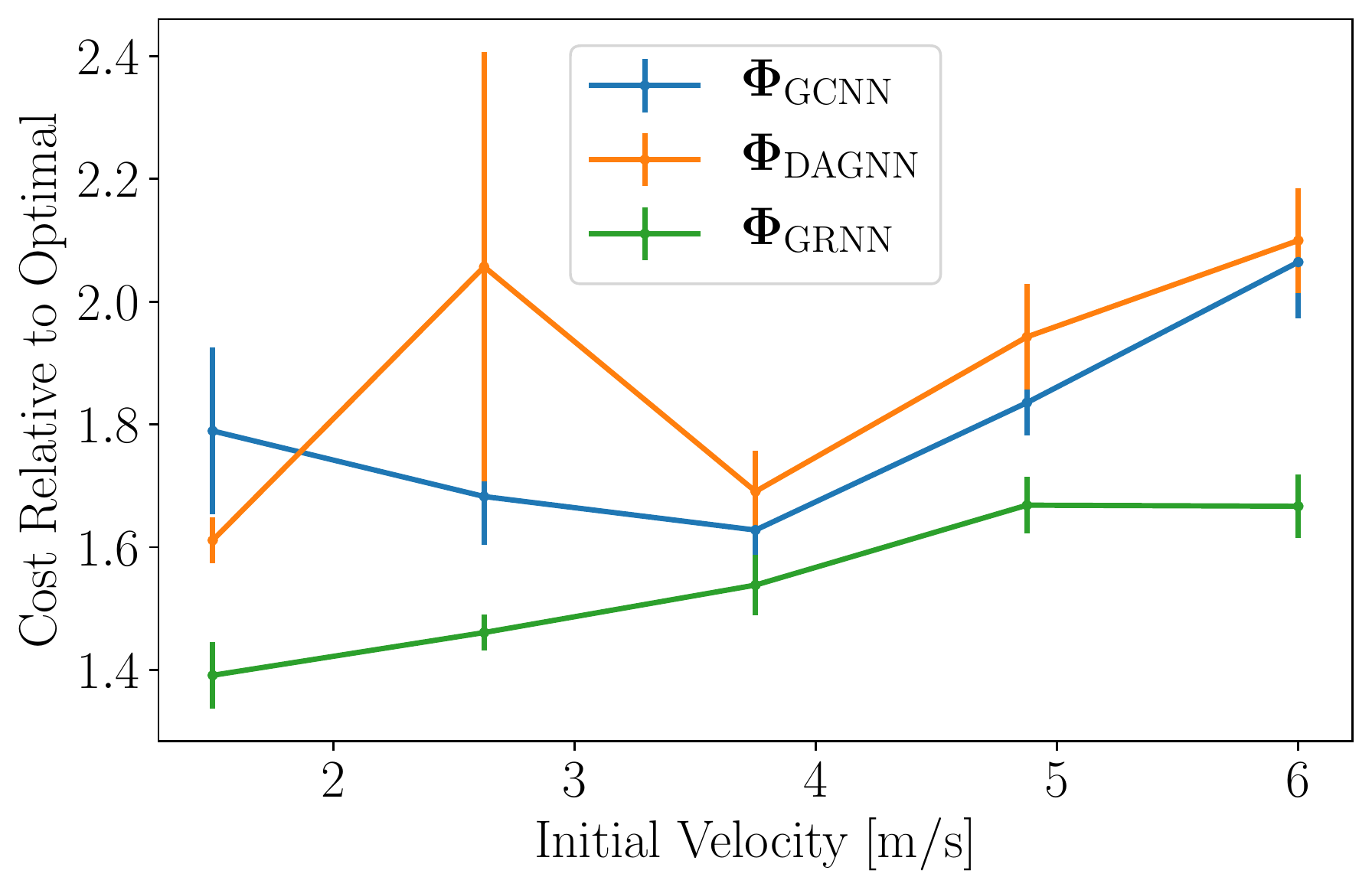}
        \caption{Initial velocity}
        \label{subfig:initVel}
    \end{subfigure}
    \hfill
    \begin{subfigure}{0.495\columnwidth}
        \centering
        \includegraphics[width = \textwidth]{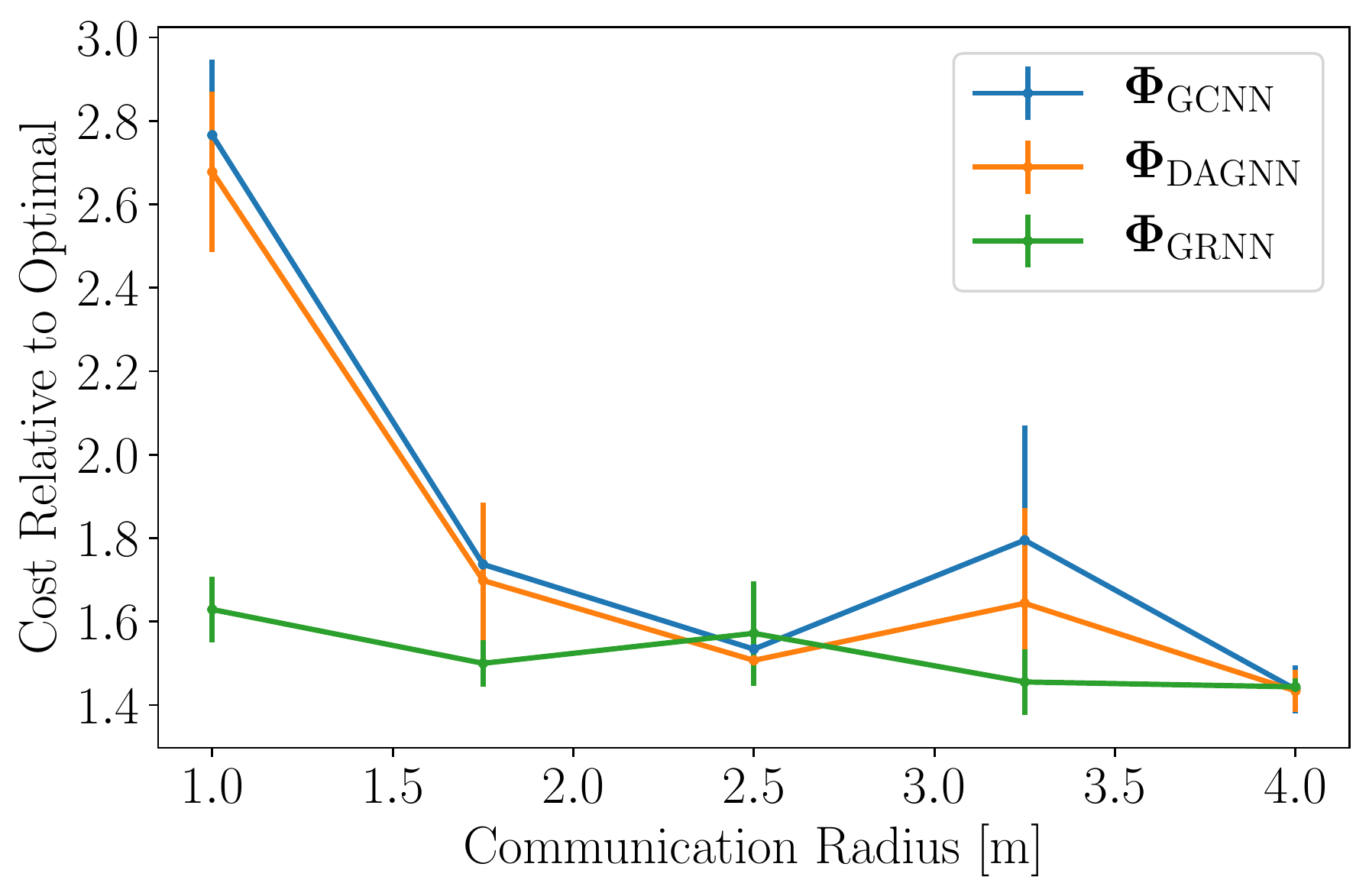}
        \caption{Communication Radius}
        \label{subfig:initRadius}
    \end{subfigure}
    \caption{Change in the cost, relative to the optimal cost, for different values of \subref{subfig:initVel} initial velocity, and \subref{subfig:initRadius} communication radius. The relative values of $\bbPhi_{\text{GC}}$ exceed $7.0(\pm 1.2)$ and $7.9(\pm 1.4)$, respectively, and thus are not shown.}
    \label{fig:init}
\end{figure}
%

\myparagraph{Experiments.} First, we test different values of features $G \in \{16, 32, 64\}$ and filter taps $K \in \{2,3,4\}$. Results are shown in Tables \ref{tab:hParamGC}-\subref{tab:hParamGRNN}. We see that the linear graph convolution $\bbPhi_{\text{GC}}$ has a performance that is five times worse than the nonlinear architectures. This is because we know that even for simple linear problems, the optimal decentralized solution is nonlinear \cite{Witsenhausen68-Counterexample}. Then we see that the $\bbPhi_{\text{GRNN}}$ exhibits the best performance, and that $\bbPhi_{\text{GCNN}}$ and $\bbPhi_{\text{DAGNN}}$ are quite similar, with the latter being slightly better. We also observe that more features $G$ improves performance in this range, but not necessarily larger $K$. From this simulation we select the best pair $(G,K)$ for each of the four architectures and keep them for the following experiments.

Second, we run tests for different initial conditions, namely different initial velocities (Fig.~ \ref{subfig:initVel}) and different communication radius (Fig.~\ref{subfig:initRadius}). These experiments test the robustness of the architectures to different initial conditions. We observe in Fig.~\ref{subfig:initVel} that larger initial velocities implies harder to control flocks, and thus the performance decreases as the initial velocities grow. Nevertheless, the GRNN seems to be more robust than the GCNN and the DAGNN. With respect to the communication radius, we observe in Fig.~\ref{subfig:initRadius} that the larger the communication radius, the easier the flock is to control. This is expected since more agents can be reached and thus information travels faster with less delay. Again, the more robust architecture is the GRNN.

As a third and final experiment, we run a test on transferring at scale. We train the architectures for $50$ agents, but then we test them on $N \in \{50, 62, 75, 87, 100\}$ agents. Results are shown in Table~\ref{tab:scalability}. We observe that all nonlinear architectures (GCNN, DAGNN and GRNN) have virtually perfect scalability, keeping the same performance as the number of agents increases. This is due to their equivariance and stability properties \cite{Gama19-Stability, Ruiz20-GRNN}. In essence, this last experiment shows that it is possible to learn a decentralized controller in a small network setting and then, once trained, transfer this solution to larger networks, successfully scaling up.


\section{Conclusions} \label{sec:conclusions}



Successful deployment of dynamical systems comprised of autonomous agents rests upon successful control. In this work, we proposed a distributed control framework using graph neural networks. While GNNs are naturally distributed and require only local information provided by nearby agents, we have adapted them to account for delays as well. By means of imitation learning, we showed how to train these models to effectively learn a decentralized controller by tracking optimal centralized trajectories. Once trained, these models can be successfully deployed in new scenarios without need of a centralized solution, thanks to their properties of permutation equivariance and stability which allow for scalability and transferability. We tested this framework in the problem of flocking, where a team of agents starts flying at random velocities, but need to coordinate in order to flock together while avoiding collisions. The experiments showcase the success of GCNNs and GRNNs in learning to flock, and in scaling up to larger teams of agents; with GRNNs exhibiting better performance than the existent DAGNNs.


\vfill\pagebreak


\bibliographystyle{bibFiles/IEEEbib}
\bibliography{bibFiles/myIEEEabrv,bibFiles/biblioFlocking}

\end{document}